\documentclass[runningheads]{llncs}
\usepackage{amsmath,amssymb,amsfonts}
\usepackage{bbm}
\usepackage{algorithmic}
\usepackage[backend=bibtex, sorting=none, style=ieee, citestyle=numeric-comp]{biblatex}
\usepackage{graphicx}
\usepackage{import}
\usepackage[ruled,vlined]{algorithm2e}
\usepackage{subcaption, booktabs}
\usepackage[usenames,dvipsnames]{xcolor}
\usepackage{caption}
\usepackage{tikz}
\usepackage{mathabx}
\usepackage{makecell}
\usepackage{arydshln}
\usepackage{physics}
\usepackage{multirow}
\usepackage[normalem]{ulem}
\usepackage{orcidlink}
\usepackage[T1]{fontenc}
\usepackage{hyperref}

\newcommand{\pembedding}{Preference Embedding\xspace}
\newcommand{\pspace}{Preference Space\xspace}
\newcommand{\mhash}{\textsc{MinHash}\xspace}
\newcommand{\rzhash}{\textsc{RuzHash}\xspace}
\newcommand{\iforest}{i\textsc{Forest}\xspace}
\newcommand{\itree}{i\textsc{Tree}\xspace}
\newcommand{\pforest}{PI-\textsc{Forest}\xspace}
\newcommand{\viforest}{PI-\textsc{Forest}\xspace}

\newcommand{\rzhitree}{\rzhash-\textsc{iTree}\xspace}
\newcommand{\rzhitrees}{\rzhash-\textsc{iTrees}\xspace}
\newcommand{\rzhiforest}{\rzhash-\textsc{iForest}\xspace}

\newcommand{\PIF}{\textcolor{orange}{\textsc{PIF}}\xspace}
\newcommand{\PIFB}{\textcolor{RoyalBlue}{\textsc{PIF-B}}\xspace}
\newcommand{\PIFR}{\textcolor{ForestGreen}{\textsc{PIF-R}}\xspace}
\newcommand{\RHF}{\textcolor{violet}{\textsc{RHF}}\xspace}
\newcommand{\RHFB}{\textcolor{red}{\textsc{RHF-B}}\xspace}

\DeclareRobustCommand{\vect}[1]{
  \ifcat#1\relax
    \boldsymbol{#1}
  \else
    \vb*{#1}
\fi}

\title{Hashing for Structure-based Anomaly Detection}

\author{
    Filippo Leveni\inst{1}\orcidlink{0009-0007-7745-5686} \and 
    Luca Magri\inst{1}\orcidlink{0000-0002-0598-8279} \and
    Cesare Alippi\inst{1,2}\orcidlink{0000-0003-3819-0025} \and 
    Giacomo Boracchi\inst{1}\orcidlink{0000-0002-1650-3054}
}
\institute{
    Politecnico di Milano (DEIB)\\
    \email{\{filippo.leveni, luca.magri, cesare.alippi, giacomo.boracchi\}@polimi.it}\\
    \and
    Università della Svizzera italiana\\
    \email{cesare.alippi@usi.ch}\\
}

\begin{document}
    \maketitle
    \begin{abstract}
       We focus on the problem of identifying samples in a set that do not conform to structured patterns represented by low-dimensional manifolds. An effective way to solve this problem is to embed data in a high dimensional space, called \pspace, where anomalies can be identified as the most isolated points. In this work, we employ Locality Sensitive Hashing to avoid explicit computation of distances in high dimensions and thus improve Anomaly Detection efficiency.
       Specifically, we present an isolation-based anomaly detection technique designed to work in the \pspace which achieves state-of-the-art performance at a lower computational cost.
       Code is publicly available at \url{https://github.com/ineveLoppiliF/Hashing-for-Structure-based-Anomaly-Detection}.
    \end{abstract}

    \section{Introduction}
        \label{sec:introduction}
        Anomaly Detection, i.e., the task of identifying anomalous instances,
        is employed in a wide range of applications such as detection of frauds in financial transactions~\cite{AhmedMahmood16}, faults  in manufacturing ~\cite{Miljkovic11},  intrusion in computer networks~\cite{LazarevicErtoz03}, risk analysis in medical data~\cite{UkilBandyoapdhyay16} and predictive maintenance~\cite{DeBenedettiLeonardiAl18}. 
        
        Most anomaly detection approaches identify anomalies as those points that lie in low density regions~\cite{ChandolaBanerjee09}.
        However, in many real word scenarios, genuine data lie on low-dimensional manifolds and, in these situations, a density analysis falls short in characterizing anomalies, that are best characterized in terms of their conformity to these low dimensional structures.
        For example, images of the face of the same subject lie on a low-dimensional subspace~\cite{BasriJacobs03}, while faces of different subjects are far away from that subspace, regardless of data density.

        In this work we focus on \emph{Structure-based Anomaly Detection}, i.e., identifying data that do not conform to any structure describing genuine data.
        Specifically, we introduce \rzhiforest, a novel anomaly detection method that is both more accurate and efficient than Preference Isolation Forest (\pforest)~\cite{LeveniMagriAl21}.
        
        \begin{figure}
            \centering
            \begin{subfigure}[t]{.325\linewidth}
                \centering
                \includegraphics[height=.8\linewidth]{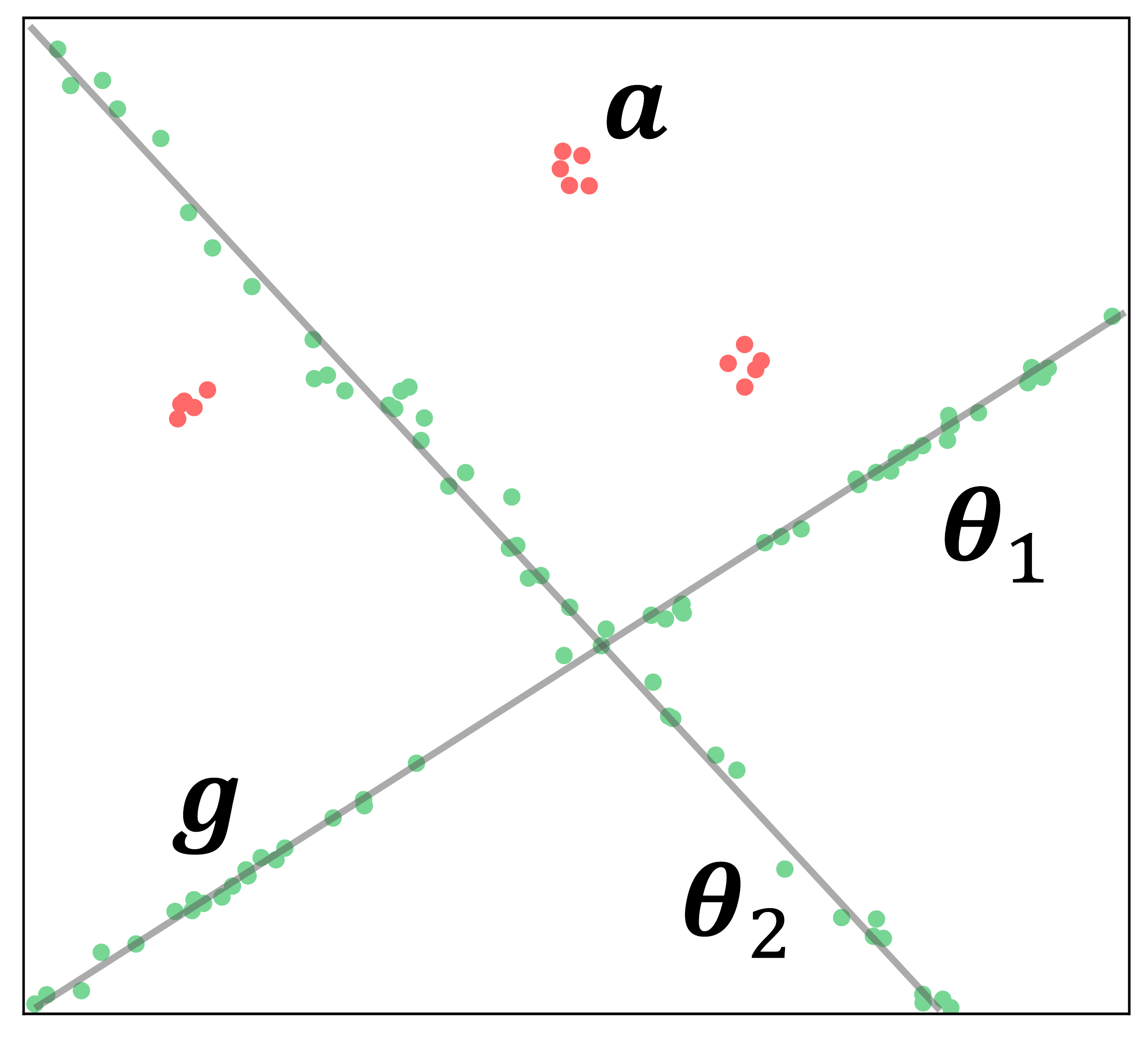}
                \caption{Input data $X = G \cup A$.}
                \label{fig:input}
            \end{subfigure}
            \hfill
            \begin{subfigure}[t]{.325\linewidth}
                \centering
                \includegraphics[height=.8\linewidth]{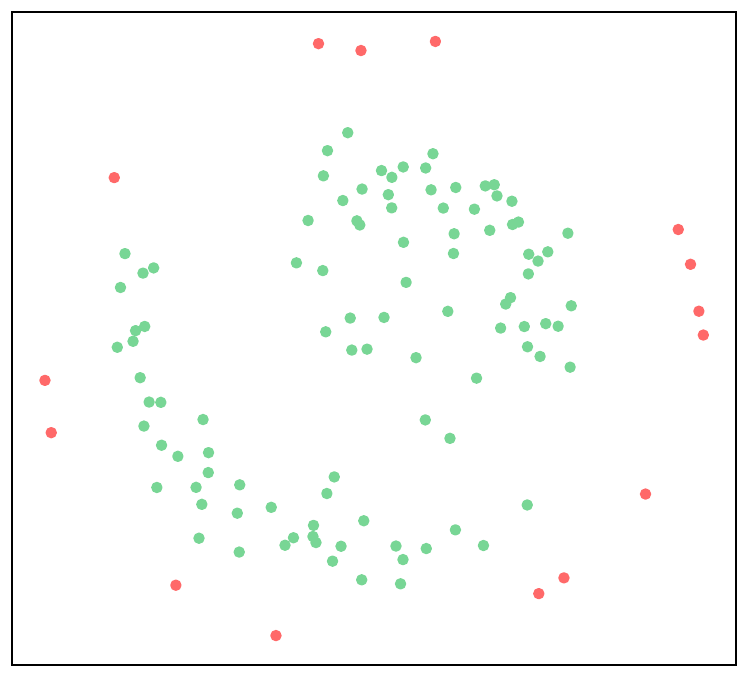}
                \caption{Preference space $\mathcal{P}$.}
               \label{fig:embedding}
            \end{subfigure}
            \hfill
            \begin{subfigure}[t]{.325\linewidth}
                \centering
                \includegraphics[height=.8\linewidth]{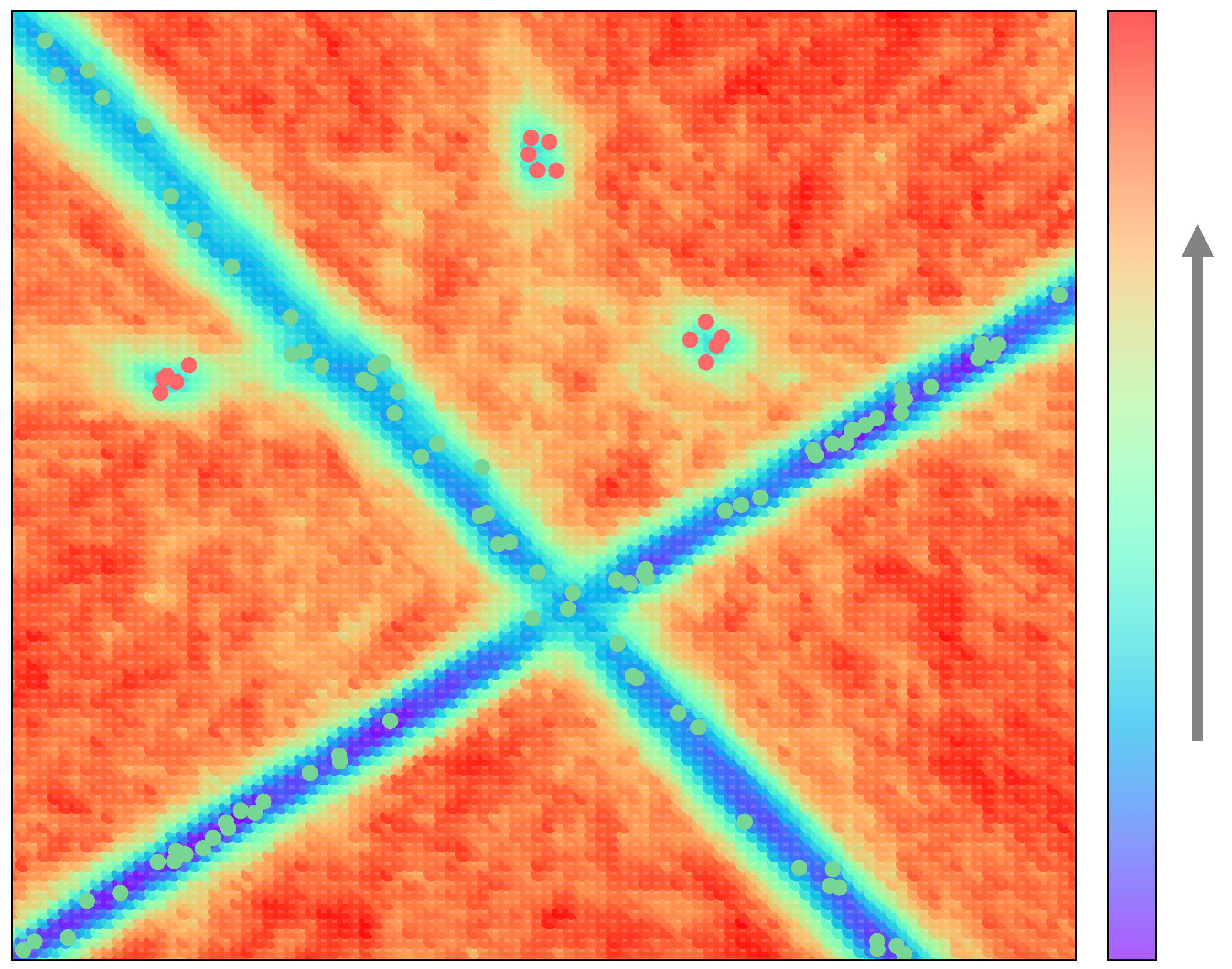}
                \caption{Anomaly scores $\alpha$.}
                \label{fig:output}
            \end{subfigure}
            \caption{\rzhiforest detects anomalies in $X$ that do not conform to structures. (a) Genuine points $G$, in green,  described by two lines of parameters $\vect{\theta}_1$ and $\vect{\theta}_2$, and anomalies $A$ in red. (b) Data are mapped to a high-dimensional \pspace where anomalies result in isolated points (visualized via MDS~\cite{Kruskal64a}). (c) Anomaly score  $\alpha(\cdot)$ (color coded) is computed via \rzhiforest.}
            \label{fig:input_output}
        \end{figure}
        In Figure~\ref{fig:input} we illustrate our method with a toy example. Genuine data $G$, depicted in green, nearly lies on 1-dimensional structures (lines $\vect{\theta}_1$ and $\vect{\theta}_2$), while anomalous data $A$ are far away from them. We embed data in a high-dimensional space (Figure~\ref{fig:embedding}), called \pspace and endowed with the Ruziska~\cite{Ruzicka58} distance, where anomalies are detected as the most isolated points.
        The explicit computation of distances in the \pspace, as done in \cite{LeveniMagriAl21}, is computational demanding, therefore we propose \rzhash, a Locality Sensitive Hashing (LSH)~\cite{GionisIndyk99} scheme that efficiently approximates distances.
        This LSH scheme is embedded in our \rzhiforest anomaly detection algorithm, and yields an anomaly score for each point.
        
        We performed experiments on both synthetic and real data to compare \rzhiforest and \pforest. Results show that our novel LSH-based approach improves \pforest both in terms of anomaly detection performance and computational time, with a speed up factor of $\times 35\%$ to $\times 70\%$.

    \section{Problem formulation}
        \label{sec:problem_formulation}
        The problem of structured anomaly detection can be framed as follows. Given a dataset  $X = G \cup A \subset \mathcal{X}$, $G$ and $A$ are the set of genuine and anomalous data respectively, and $\mathcal{X}$ is the ambient space.
        We assume that genuine data $\vect{g} \in G$ are close to the solutions of a parametric equation $\mathcal{F}(\vect{g}, \vect{\theta})=0$, that depends on an unknown vector of parameters $\vect{\theta}$. For example, in Figure~\ref{fig:input} genuine structures are described by lines, thus $\mathcal{F}(\vect{g}, \vect{\theta}) = \theta_1 g_1 + \theta_2 g_2 + \theta_3$ and, due to noise, genuine data satisfy the equation only up to a tolerance $\epsilon > 0$, namely $|\mathcal{F}(\vect{g}, \vect{\theta})| < \epsilon$.
        Moreover, genuine data may be described by multiple models $\{\vect{\theta}_i\}_{i = 1, \dots, k}$ whose number is typically unknown.
        In contrast, anomalous data $\vect{a} \in A$ are far from satisfying the parametric equation of any model instance  and $\mathcal{F}(\vect{a}, \vect{\theta}_i) \gg \epsilon \ \forall \vect{\theta}_i$.

        Structured-based Anomaly Detection aims to produce an anomaly scoring function $\alpha: X \rightarrow \mathbb{R}^+$ such that $\alpha(\vect{a}) \gg \alpha(\vect{g})$, as depicted in Figure~\ref{fig:output} where higher scores are in red and lower scores in blue.

    \section{Related Work}
        \label{sec:related_works}
        Among the wide literature on anomaly detection~\cite{ChandolaBanerjeeAl09}, we focus on isolation-based methods, as they reach state-of-the-art performance at low computational and memory requirements.    
        Isolation-based approaches can be traced back to  Isolation Forest ~\cite{LiuTing12} (\iforest), where anomalies are separated by building a forest of randomly generated binary trees (\itree) that recursively partition the data by axis-parallel splits.
        The number of splits required to isolate any point from the others is inversely related to its probability of being anomalous. In other words, anomalies are more likely to be separated in the early splits of the tree and result in shorter paths. Thus, the average  path lengths, computed with respect to a forest of random trees,  translates into a reliable anomaly score.
        Several improvements over the original \iforest framework have been introduced.
        Extended Isolation Forest~\cite{HaririKindAl19} and Generalized Isolation Forest~\cite{LesoupleBaudoinAl21} overcome the limitation of axis-parallel splits, while Functional Isolation Forest~\cite{StaermanMozharovskyiAl19} extends \iforest beyond the concept of point-anomaly, to identify functional-anomalies.
        The connection between \iforest and Locality Sensitive Hashing is investigated in \cite{ZhangDouAl17}. In particular, the splitting process of an \itree is interpreted as a Locality Sensitive Hashing (LSH) of the $\ell_1$ distance, where points that are nearby according to $\ell_1$ are assigned to the same bucket. LSH schemes allow to strike a good trade-off between effectiveness and efficiency, but are limited to identify density-based anomalies.
        
        The literature on structure-based anomaly detection is less explored than its density-based counterpart. The pioneering work was \cite{LeveniMagriAl21}, where \pforest, a variant of \iforest, is presented to detect structured anomalies.
        \pforest consists to randomly sample a set of low-dimensional structures from data points and to embed data into an high-dimensional space, called \pspace, where each point is described in terms of its adherence to the sampled structures. Here, \pforest identifies anomalies as the most isolated points according to the Jaccard or Tanimoto distance.
        Specifically, \pforest leverages on nested Voronoi tessellations built in a recursive way in the \pspace to instantiate isolation-trees. Although this approach effectively isolates anomalous data, building Voronoi tessellations carries the computational burden of explicitly computing distances in high dimensions, impacting negatively on the computational performance.
        
        In this work, we address this problem by presenting a novel LSH scheme to approximate distances in the \pspace. The problem of speeding up computation of distances in the \pspace has been addressed in~\cite{magri2018reconstruction}, where \mhash has been used to cluster points according to the Jaccard distance. However, the focus of~\cite{magri2018reconstruction} was to identify structures rather than anomalies and it is limited to deal with binary preferences.

    \section{Method}
        \label{sec:method}
        \begin{algorithm}[t]
            \caption{\rzhiforest \label{alg:main}}
            \DontPrintSemicolon
            \SetNoFillComment
            \KwIn{$X$ - input data, $t$ - number of trees, $\psi$ - sub-sampling size, $b$ - branching factor}
            \KwOut{Anomaly scores $\{\alpha(\vect{x}_j)\}_{j=1,\ldots,n}$}
            \begin{small}
                \tcc{\pembedding\!\!\!}
            \end{small}
            Sample $m$ models $\{\vect{\theta}_i\}_{i=1,\ldots,m}$ from $X$ \label{line:begin_embedding}\\
            $P \leftarrow \{\vect{p}_j \,|\, \vect{p}_j = \mathcal{E}(\vect{x}_j)\}_{j=1,\ldots,n}$ \label{line:end_embedding}\\
            \begin{small}
                \tcc{\rzhiforest\!\!\!}
            \end{small}
            $F \leftarrow \emptyset$ \label{line:begin_forest_construction}\\
            \For{$k = 1$ \normalfont{to} $t$}
                {$P_\psi \leftarrow \text{\textsc{Subsample}}(P, \psi)$ \label{line:subsample} \\
                 $T_k \leftarrow \text{\rzhitree}(P_\psi, b)$ \label{line:tree_construction} \\
                 $F \leftarrow F \cup T_k$ \label{line:end_forest_construction}}
            \begin{small}
                \tcc{Anomaly score computation}
            \end{small}
            \For{$j = 1$ \normalfont{to} $n$ \label{line:begin_detection}}
                {
                \For{$k = 1$ \normalfont{to} $t$}
                     {$\vect{p}_j \leftarrow \text{$j$-th point in $P$}$ \label{line:point} \\
                      $T_k \leftarrow \text{$k$-th \rzhitree in $F$}$ \label{line:tree} \\
                      ${h}_k(\vect{p_j}) \leftarrow \text{\textsc{Height}}(\vect{p}_j, T_k)$ \label{line:height}}
                $\alpha(\vect{x}_j) \leftarrow $ Eq.\eqref{eq:anomaly_score} \label{line:anomaly_score} 
                }
            return $\{\alpha(\vect{x}_j)\}_{j=1,\ldots,n}$ \label{line:end_detection} \\
        \end{algorithm}

        The proposed method, summarized in Algorithm \ref{alg:main}, is composed of two main steps. In the first one (lines~\ref{line:begin_embedding}-\ref{line:end_embedding}), we map input data $X \subset \mathcal{X}$ to $P \subset \mathcal{P}$ via the \pembedding $\mathcal{E}(\cdot)$, where $\mathcal{P}$ is an high-dimensional \pspace. The mapping process $\mathcal{E}(\cdot)$ follows closely the \pembedding of \cite{LeveniMagriAl21}, and it is reported in Section~\ref{subsec:preference_embedding}.
        The major differences with respect to \cite{LeveniMagriAl21} reside in the second step, and in particular in the computation of distances in the \pspace. First, we do not use the Tanimoto distance~\cite{Lipkus99}, but rather we introduce the Ruzicka distance~\cite{Ruzicka58}, which demonstrates comparable isolation capabilities. Secondly, we define a novel Locality Sensitive Hashing scheme, called \rzhash to efficiently approximate Ruzicka distances avoiding their explicit computation.
        Specifically, we build an ensemble of isolation trees termed \rzhiforest (lines~\ref{line:begin_forest_construction}-\ref{line:end_forest_construction}) as described in Section \ref{subsec:hashing_forest}. Each \rzhitree of the forest recursively splits the points based on our \rzhash Local Sensitive Hashing as detailed in Section \ref{subsec:ruzhash_itree}. This separation mechanism produces an anomaly score $\alpha(\vect{x})$ for each $\vect{x} \in X$ (lines \ref{line:begin_detection} -\ref{line:end_detection}), as discussed in Section~\ref{subsec:ruzhash}.
        
        \subsection{Preference Embedding}
            \label{subsec:preference_embedding}
            \pembedding has been widely used in the multi-model fitting literature~\cite{ToldoFusiello08,MagriFusiello14,MagriLeveniAl21} and then employed in Structure-based Anomaly Detection~\cite{LeveniMagriAl21}.
            \pembedding consists in a mapping $\mathcal{E}\colon \mathcal{X} \to \mathcal{P}$, from the ambient space $\mathcal{X}$ to the \pspace $\mathcal{P} = [0, 1]^m$.
            Such mapping is obtained by sampling a pool $\{\vect{\theta}_i\}_{i=1,\ldots,m}$ of $m$ models from the data $X$ using a RanSaC-like strategy~\cite{FischlerBolles81} (line~\ref{line:begin_embedding}): the minimal sample sets -- containing the minimum number of points necessary to constrain a parametric model -- are randomly sampled from the data to determine models parameters.
            Then (line~\ref{line:end_embedding}), each sample $\vect{x} \in \mathcal{X}$ is embedded to a vector $\vect{p} = \mathcal{E}(\vect{x}) \in \mathcal{P}$ whose $i$-th component is defined as:
            \begin{equation}
                \label{eq:preference}
                p_i  =
                \begin{cases}
                    \phi(\delta_{i}) &\text{if $|\delta_{i}| \leq \epsilon$ }\\
                                    0 &\text{otherwise}
                \end{cases},
            \end{equation}
            where $\delta_i = \mathcal{F}(\vect{x},\vect{\theta}_i)$, measures the residuals of $\vect{x}$ with respect to model $\vect{\theta}_i$ and $\epsilon = k\sigma$ defines an inlier threshold proportional to the standard deviation $\sigma$ of the noise. The preference function $\phi$ is then defined as:
            \begin{equation}
                \label{eq:pref}
                \phi(\delta_i) = e^{-\frac{1}{2}(\frac{\delta_i}{\sigma})^2}.
            \end{equation}
            We explored also a different definition of $\phi$, namely the \emph{binary preference} function that is $\phi(\delta_i) = 1$ when $|\delta_i| \leq \epsilon$ and $0$ otherwise. Hereinafter, we will refer to $\mathcal{P} = \{0, 1\}^m$ as the \emph{binary preference space} to distinguish it from  the \emph{continuous} one $\mathcal{P} =[0,1]^m$.

        \subsection{\rzhiforest}
            \label{subsec:hashing_forest}
            We perform anomaly detection in the \pspace exploiting a forest of isolation trees similarly to \cite{LeveniMagriAl21} (lines~\ref{line:begin_forest_construction}-\ref{line:end_detection}). The fundamental difference of our \rzhiforest  with respect to \cite{LeveniMagriAl21} is that our ensemble of \rzhitrees  bypass the distance computation in the preference space.
            \\
            We identify two main steps: the \emph{training} of \rzhiforest $F = \{T_k\}_{k=1}^t$ (lines \ref{line:begin_forest_construction}-\ref{line:end_forest_construction}) and the \emph{testing} of vectors $P$ via every \rzhitree $T_k \in F$ (lines \ref{line:begin_detection}-\ref{line:end_detection}).
            As regard the training, we build every \rzhitree on a different subset $P_\psi \subset P$ of $\psi$ vectors sampled from $P$ (line~\ref{line:subsample}).
            During testing, for every point $\vect{p}_j \in P$ (line~\ref{line:point}) and every tree  $T_k \in F$ in the forest (line~\ref{line:tree}), we compute the heights $h_k(\vect{p}_j)$ reached in $T_k$ by $\vect{p}_j$ (line~\ref{line:height}).
            The main intuition is that each $T_k$ returns, on average, noticeable smaller heights for anomalies than for genuine points, since isolated points are more likely to be separated early in the recursive splitting process.
            Thus, the heights are collected in a vector $\vect{h(\vect{p}_j)} = [h_1(\vect{p}_j), \dots, h_t(\vect{p}_j)]$ and the anomaly score $\alpha(\cdot)$ is computed  (line \ref{line:anomaly_score}) as:
            \begin{equation}
                \label{eq:anomaly_score}
               \alpha(\vect{x}_j) = 2^{-\frac{E(\vect{h}(\vect{p}_j))}{c(\psi)}},
            \end{equation}
            where $E(\vect{h}(\vect{p}_j))$ is the mean over the elements of $\vect{h}(\vect{p}_j)$ and $c(\psi) = \log_{b}\psi$ is an adjustment factor as a function of the tree subsampling size $\psi$. Our method is agnostic with respect to the specific choice of anomaly score, and other techniques, as discussed in \cite{MensiBicego21}, can be  employed as well.

       \subsection{\rzhitree}
            \label{subsec:ruzhash_itree}
            The  construction of each  \rzhitree $T_k$ (line~\ref{line:tree_construction}) is detailed as follow.
            We are given a subset $P_\psi$, with cardinality $\psi$, uniformly sampled from all the input points $P$ embedded in the \pspace (line~\ref{line:subsample}) and a branching factor $b \in \{1, \ldots, m\}$.
            At each node, $P_\psi$ is splitted in $b$ branches using \rzhash. 
            This scheme is recursively executed until either: (\emph{i}) the current node contains a number of points less than $m$ or (\emph{ii}) the tree reaches a maximum height, set by default at $\log_m\psi$ (an approximation for the average tree height \cite{Knuth98}).
            \begin{figure}[t]
                    \centering
                    \includegraphics[width=0.85\linewidth]{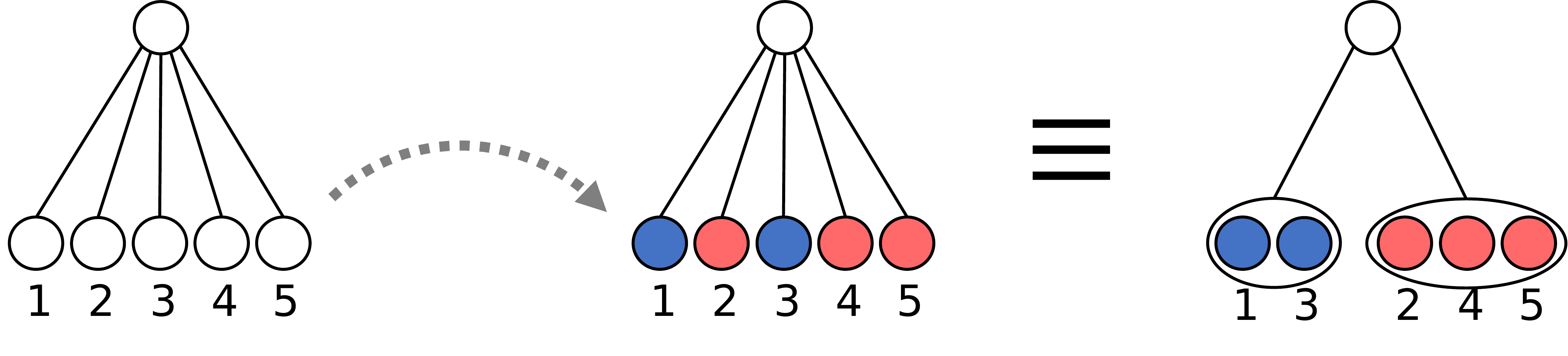}
                    \caption{On the left, split performed by \rzhash. In the middle, nodes aggregation where the groups has been color coded. On the right, resulting tree with branching factor $b = 2$.}
                    \label{fig:partitioning}
            \end{figure} 
           
            \rzhash is designed to split the data in $m$ leaves, where $m$ equals the dimension of $\mathcal{P}$.  However, we experienced that lower branching factors resulted in slightly better performance. Therefore, we accommodate for a different branching factor $b$ by randomly aggregating in $b < m$ groups the nodes produced by \rzhash after each split of the tree.
            Figure~\ref{fig:partitioning} shows an example of aggregation process when $m = 5$ and $b = 2$. We can see on the left $m$ leaves performed by \rzhash, and in the middle the aggregation of the nodes where the groups have been color coded. On the right, we can see the resulting tree with branching factor $b = 2$.
            The branching factor controls the average tree height. Therefore, the maximum tree height becomes $\log_b\psi$.

         \subsection{\rzhash}
            \label{subsec:ruzhash}
            Instead of leveraging explicitly on distances, the splitting procedure implemented in each node of \rzhitree is based on \rzhash, our novel Locality Sensitive Hashing process, designed to approximate the Ruziska distance.
            In this way, we greatly reduce the computational burden of the method.
            
            We consider Ruzicka instead of Tanimoto to measure distances in the preference space. This is due to the fact that has not yet been proven whether a hashing scheme for Tanimoto could even exist. Moreover, our experiments demonstrate that Ruzicka and Tanimoto distances achieves comparable performance in isolating anomalous points in the \pspace.
            \\
            Given two preferences vectors $\vect{p}, \vect{q} \in \mathcal{P}$, their Ruzicka distance is defined as:
            \begin{equation}
                \label{eq:ruzicka_vect}
                R(\vect{p}, \vect{q}) = 1 - \frac{\sum_{i=1}^{m} \min(p_i, q_i)}{\sum_{i=1}^{m} \max(p_i, q_i)}.
            \end{equation}
            In practice, the higher the preferences granted to the same models $\{\vect{\theta}_i\}_{i = 1, \dots, m}$, represented by components $p_i$ and $q_i$, the closer the vectors $\vect{p}$ and $\vect{q}$ are.
            \begin{figure}[tb]
                \hspace{0.3cm}
                \begin{minipage}[b]{.45\linewidth}
                    \centering
                    \captionsetup{width=.9\linewidth}
                    \includegraphics[width=\linewidth]{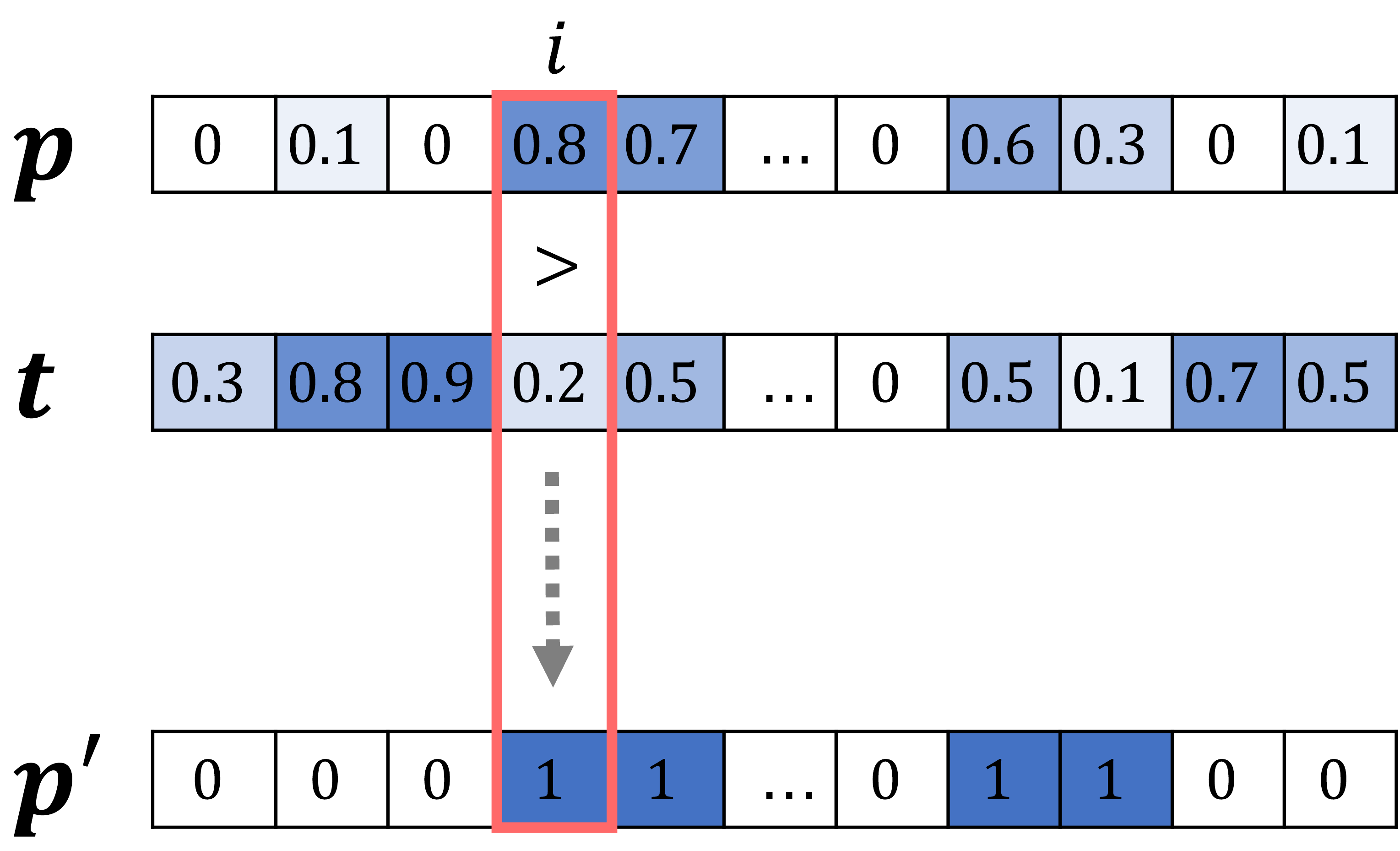}
                    \caption{Example of binarization.}
                    \label{fig:thresholding}
                \end{minipage}
                \hspace{0.5cm}
                \begin{minipage}[b]{.45\linewidth}
                    \centering
                    \captionsetup{width=.8\linewidth}
                    \includegraphics[height=.65\linewidth]{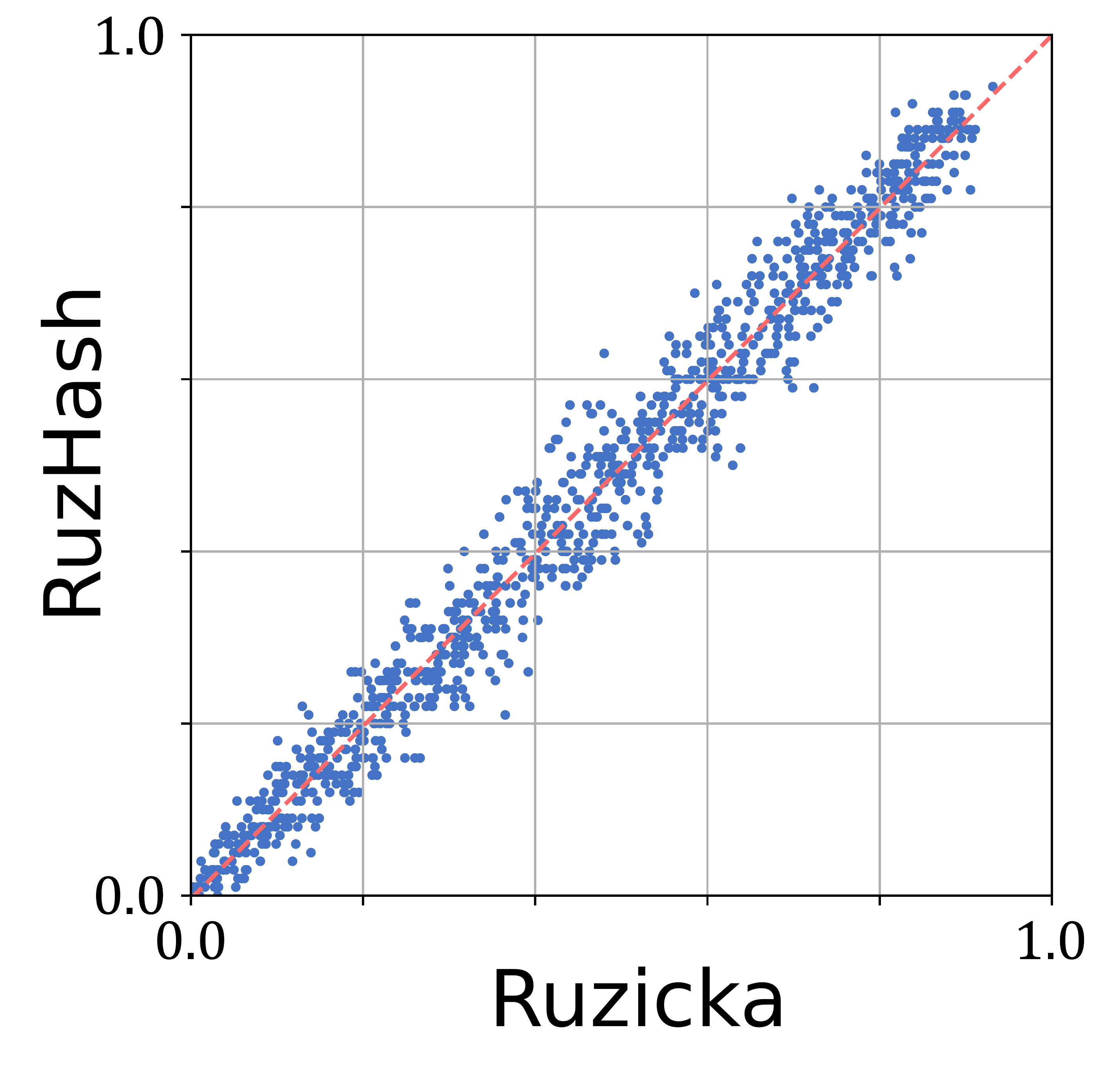}
                    \vspace{-0.525cm}
                    \caption{Correlation between Ruzicka and \rzhash.}
                    \label{fig:correlation}
                \end{minipage}
            \end{figure}
            
            Our locality sensitive hashing scheme, \rzhash, approximates the Ruzicka distance in the \pspace as follows.
            First, vectors $\vect{p} \in \mathcal{P} = [0, 1]^m$ are binarized yielding points $\vect{p}' \in \{0, 1\}^m$. The binarization is performed by a component-wise comparison of $\vect{p}$ with a randomly sampled vector $\vect{t} \in [0, 1]^m$, ad depicted in Figure~\ref{fig:thresholding}. Formally, we define a vector $\vect{t} = [t_1, \dots, t_m]$ as the realization of a multivariate random vector $\boldsymbol{T} = [T_1, \dots, T_m]$, where $T_i$ is a uniform random variable $T_i \sim \mathcal{U}_{[0, 1)}$, and use it to binarize \emph{all} points $\vect{p} \in [0, 1]^m$. If the $i$-th component $p_i$ of a point $\vect{p}$ is greater than $t_i$, the binarized component equals $1$, otherwise it is set to $0$. In formulae, the component-wise binarization procedure is defined as:
            \begin{equation}
                p'_i =
                \begin{cases}
                    1 & \text{if $p_i > t_i$}\\
                    0 & \text{otherwise}
                \end{cases}.
                \label{eq:thresholding}
            \end{equation}
            Notice that, since $\vect{t}$ is a realization of random variable, and the binarization procedure \eqref{eq:thresholding} depends on $\vect{t}$, thus $\vect{p}'$ is a realization of a random variable.
            Now, given the $i$-th binarized components $p'_i, q'_i$ of two preference vectors $\vect{p}, \vect{q}$, the probability that both $p'_i$ and $q'_i$ equals $1$, is given by:
            \begin{equation}
            P(p'_i = 1 \land q'_i = 1) = P(p_i > t_i \land q_i > t_i) = \min\{p_i, q_i\}.
            \end{equation}
            Similarly, the probability that at least one of the binarized component equals $1$, is given by $P(p_i'=1 \lor q_i' =1) = \max\{p_i,q_i\}$.
            Note that these two terms are exactly those that appear in \eqref{eq:ruzicka_vect}  in the definition of the Ruziska distance  at the numerator and denominator, respectively.
            Thus, we can rewrite the Ruzicka distance in terms of the previous probabilities between binary vectors:
            \begin{equation}
            R(\vect{p}, \vect{q})= 1 - \frac{\sum_{i=1}^{m} P(p'_i =1 \land q'_i =1)}{\sum_{i=1}^{m} P(p'_i =1 \lor q'_i =1)}.
            \label{eq:ruzicka_prob}
            \end{equation}
            The Ruzicka distance is estimated by computing the value of \eqref{eq:ruzicka_prob} for different realizations of vectors $\vect{p}', \vect{q}'$, sampling several vectors $\vect{t}$.
            In this way, the problem boils down to counting the number of $1$ on the same component $i$ over the total number of non zero entries of $\vect{p}', \vect{q}'$, that is their Jaccard distance. This quantity, in turn, can be efficiently estimated using \mhash~\cite{BroderCharikar98}. The estimated value of Ruzicka converges to its theoretical value \eqref{eq:ruzicka_vect} as the number of sampled $\vect{t}$ increases. Figure~\ref{fig:correlation} depicts the nearly exact correlation existing between Ruzicka and \rzhash, computed on pairs of randomly sampled vectors such that the distances between them were uniform with respect to the Ruzicka distance.
            
            In practice, a different $\vect{t}$ is sampled at each split of every \rzhitree in the forest during the training phase. It follows that, the more splits are performed, the higher the probability that only samples close with respect to the Ruzicka distance fall in the same node and, at the same time, the more isolated points are separated from the others in the early splits.
            
            To conclude, it is worth to notice that Ruzicka distance, as the Tanimoto one, is a generalization of the Jaccard distance when preferences are in $\{0, 1\}^m$.
            In this case, $\vect{p}= \vect{p}'$ and \rzhash specializes exactly to \mhash.

        \subsection{Comparison between \iforest, \pforest and \rzhiforest}
            Table~\ref{tab:differences} summarizes the main differences between \iforest, \pforest and \rzhiforest.
            As regard the splitting scheme, \rzhiforest exploits LSH as done in \iforest. This can be appreciated from the lower computational complexity compared to the \pforest which builds Voronoi tessellations, and require the explicit computation of distances to the $b$ tessellation centers. As regard the distance employed, \rzhiforest exploits the Ruziska distance that is tailored for the \pspace, rather than relying on the $\ell_1$ distance.  Moreover, if we compare the computational complexities of \iforest and \rzhiforest, we have a speed up given by the higher branching factor of \rzhiforest compared to the fixed branching factor $b = 2$ of \iforest.
            
            \begin{table}[t]
                \centering
                \resizebox{0.92\textwidth}{!}{
                \begin{tabular}{c||c|c|c|c}
                                & \multirow{2}{*}{Splitting scheme} & \multirow{2}{*}{Distance} & \multicolumn{2}{c}{Computational complexity}                                               \\ \cline{4-5}
                                &                                   &                           & Training                                      & Testing                                    \\ \hline
                    \pforest    & Voronoi                           & Tanimoto                  & $O(\psi \: t \: b \: \log_{b} \psi)$ & $O(n \: t \: b \: \log_{b} \psi)$ \\
                    \iforest    & LSH                               & $\ell_1$                  & $O(\psi \: t \: \log_{2} \psi)$         & $O(n \: t \: \log_{2}\psi)$          \\
                    \rzhiforest & LSH                               & Ruzicka                   & $O(\psi \: t \: \log_{b} \psi)$         & $O(n \: t \: \log_{b} \psi)$         \\
                \end{tabular}
                }
                \caption{Differences between \iforest, \pforest and \rzhiforest.}
                \label{tab:differences}
            \end{table}

    \section{Experimental Validation}
        \label{sec:experimental_validation}
        In this section we evaluate the benefits of our approach for structured anomaly detection on both simulated and real datasets. In particular, we compare the performance of \rzhiforest and \pforest both in the \emph{continuous} and \emph{binary} preference space. Results show that \rzhiforest performs better in terms of both AUC and execution time.
        \subsection{Datasets}
            We consider synthetic datasets and real data employed in \cite{LeveniMagriAl21}. Synthetic datasets consist of 2D points where genuine data $G$ live along parametric structures (lines and circles) and anomalies are uniformly sampled within the range of $G$ such that $\frac{|A|}{|X|} = 0.5$.
            
            We consider a real dataset, the AdelaideRMF dataset~\cite{WongChinAl11}, that consists stereo images with annotated matching points and anomalies $A$ correspond to mismatches. The first 19 sequences refer to static scenes containing several planes, each giving rise matches described by an homography. The remaining 19 sequences are dynamic with several objects independently moving and give rise to a set of matches described by different fundamental matrices.

        \subsection{Competing methods}
            We compare \rzhiforest against \viforest~\cite{LeveniMagriAl21}, both constructed with binary or continuous \pspace. When preferences are continuous, we use the shorthand \RHF and \PIF respectively, and we indicate by \RHFB and \PIFB the two competitors when preferences are binary. In order to assess the benefits of Ruziska distance we also considered a version of \cite{LeveniMagriAl21} equipped with Ruziska instead of Tanimoto and denoted this by \PIFR. 
            
            Preferences are computed with respect to a pool of $m=10|X|$ model instances, corresponding to the genuine structures.
            The inlier threshold $\epsilon$ in \eqref{eq:pref} has been tuned as follows: we first estimate the standard deviation $\sigma$ of the noise from the data given their ground truth labels, then we fix $\epsilon = k\sigma$ where we choose $k$ to maximize the performance for both \RHF and \PIF. In particular, we set $k = 3$ for synthetic data, $k = 0.25$ for fundamental matrix and $k = 5$ for homography dataset respectively.

            We tested \viforest and \rzhiforest at the same parameters condition: number of trees in the ensemble $t = 100$, subsampling size to build each tree $\psi = 256$ while the branching factor vary in $b = [2, 4, 8, 16, 32, 64, 128, 256]$.

       \subsection{Results}
            Figure~\ref{fig:results} shows the aggregated results of our experiments. In particular, in Figure~\ref{fig:roc_auc} and \ref{fig:test_time} we show respectively the average ROC AUC and test time for all the methods at various branching factors. We produced the curves by first averaging the results of $5$ executions on synthetic and real datasets separately. We then averaged these results to get the final curves.
            
            The rundown of these experiments is that our \rzhiforest achieves higher ROC AUC values in both \RHF and \RHFB configurations. More interestingly, our method is the most stable with respect to the choice of the branching factor $b$.  \rzhiforest attains accurate results also for small values of $b$, because the tendency to underestimate the distances of the splitting procedure is compensated by the overestimation  due to the greater height of the trees. On the other hand, \PIF, \PIFB and \PIFR have a consistent performance loss when the branching factor increases. This can be ascribed to the Voronoi tessellations that enforce each node to contain at least one point. Thus, when $b\geq32$, trees are constrained to a single level, 
            and the anomaly score does no longer depend on the height of the tree but is fully determined by the adjustment factor alone ~\cite{LiuTing12, LeveniMagriAl21}, resulting in a degradation of the performances.
            These trends are confirmed in the average ROC AUC curves computed separately for synthetic and real datasets, which have not been reported due to space limitations.
            Furthermore, AUC curves of \PIF and \PIFR show that there is not a clear advantage in using the Ruziska distance over the Tanimoto one in the \pforest framework, confirming that the main advantage of our method is due to the LSH scheme and not to the different distance measures involved. 
            The gain in terms of test time is very evident in Figure~\ref{fig:test_time} and it is consistent with the computational complexities showed in Table~\ref{tab:differences}. In fact, when the branching factor $b$ increases, the test time for \rzhiforest decreases according to $\log_{b} \psi$ in all configurations, while for \viforest increases according to $b \log_{b} \psi$.

            A different visualization of the results is presented in Figure~\ref{fig:best_roc_auc_time}, where the relation between the best average ROC AUC value and the corresponding test time for each method is shown. We identify for each method the branching factor that maximizes the average ROC AUC (Figure~\ref{fig:roc_auc}) and use it to collect the corresponding test time (Figure~\ref{fig:test_time}). Dashed lines relate \rzhiforest and \viforest results that refer to the same underlying distance measure (Ruziska for the continuous case, Jaccard for the binary). It can be appreciated that \rzhiforest achieves results that are as accurate as their \pforest counterpart, but with a consistent gain in execution time. Specifically, \RHFB is $\times 35\%$ faster than \PIFB, while \RHF is $\times 70\%$ faster than \PIFR. Furthermore, \RHF is at least $\times 70\%$ faster than \PIF, which employs the Tanimoto distance.
            
            \begin{figure}[t]
               \centering
               \includegraphics[width=\linewidth]{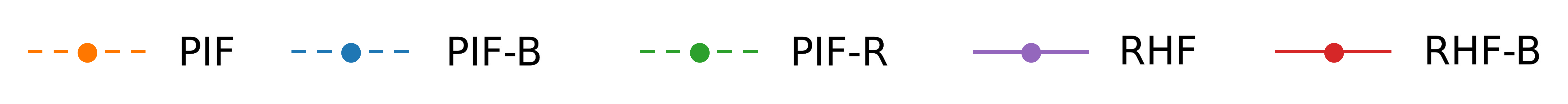}
                \\
                \hspace*{-1.9cm}
                \captionsetup[subfigure]{oneside, margin={0.75cm,0cm}}
                \subfloat[\label{fig:roc_auc}]{
                    \includegraphics[width=.4\linewidth]{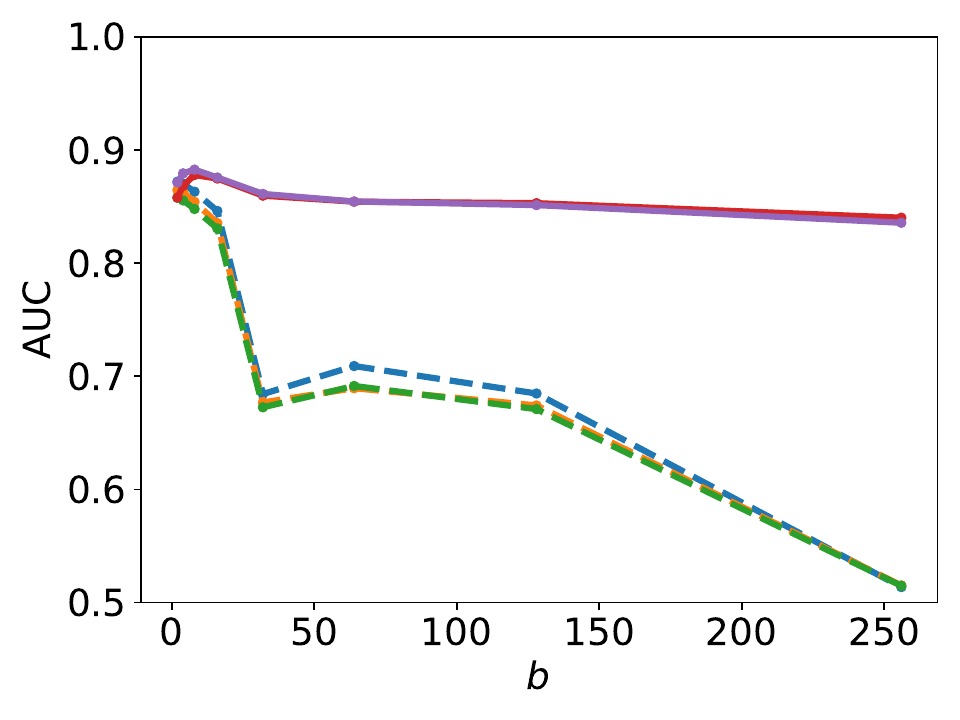}
                    \vspace{-0.25cm}
                }
                \subfloat[\label{fig:test_time}]{
                    \includegraphics[width=.4\linewidth]{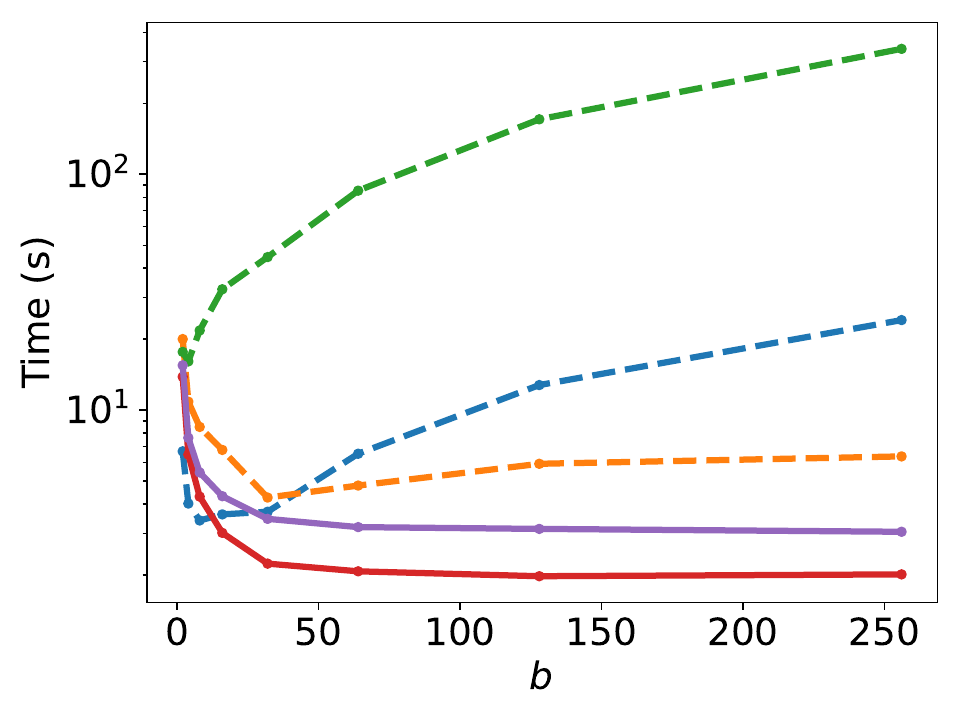}
                    \vspace{-0.25cm}
                }
                \subfloat[\label{fig:best_roc_auc_time}]{
                    \includegraphics[width=.4\linewidth]{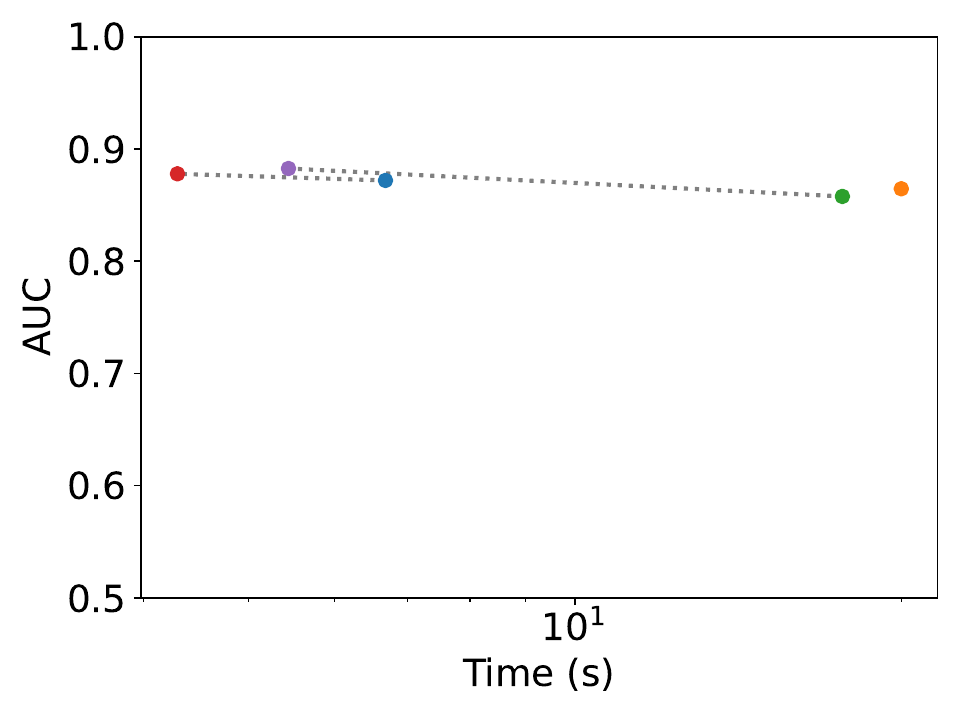}
                    \vspace{-0.25cm}
                }
                \caption{(a) Average ROC AUCs. (b) Average test times. (c) Relation between best average ROC AUC and corresponding test time.}
                \label{fig:results}
            \end{figure}
    
    \section{Conclusion and Future Directions}
        \label{sec:conclusion_and_future_directions}
        We proposed \rzhiforest, an efficient algorithm specifically designed to perform Structure-based Anomaly Detection. \rzhiforest is an isolation-based anomaly detection algorithm that works in the \pspace, whose peculiarity resides in the splitting criteria. In particular, a novel Locality Sensitive Hashing, called \rzhash, has been employed to detect the most isolated points in the \pspace with respect to the Ruzicka distance. Remarkably, \rzhash is a generalization of \mhash to the case where the considered points lie in the continuous space $\mathcal{P} = [0, 1]^m$.

        Our empirical evaluation demonstrated that \rzhiforest gain efficiency over current state-of-the-art-solutions, and has stable accuracy along different branching factors, both on synthetic and real data. A possible future direction could be to investigate others effective distances for anomaly detection in the \pspace and to define their corresponding LSH to employ in an isolation-based forest.

    \printbibliography
\end{document}